\PassOptionsToPackage{table}{xcolor}
\documentclass[runningheads]{llncs}

\usepackage{eccv}

\usepackage{eccvabbrv}

\usepackage{graphicx}
\usepackage{booktabs}

\usepackage[accsupp]{axessibility}  

\usepackage{hyperref}

\usepackage{orcidlink}

\usepackage{multirow}
\usepackage{makecell}
\usepackage{tikz}
\usetikzlibrary{shapes.geometric}
\usetikzlibrary{fit}
\usetikzlibrary{decorations.pathreplacing}
\usetikzlibrary{positioning, shapes, calc, arrows, arrows.meta, backgrounds, shadows, matrix}
\usepackage{arydshln} 
\usepackage{booktabs}
\usepackage{pifont}
\usepackage{algorithm}
\usepackage{algpseudocode}
\usepackage{pgfplots}
\pgfplotsset{compat=1.17}
\usepackage{bm} 
\usepackage{fontawesome5} 
\usepackage{adjustbox}
\definecolor{lightgray}{gray}{0.95}
\definecolor{color_man}{RGB}{151, 178, 228}
\definecolor{color_sk}{RGB}{175, 248, 204}
\definecolor{color_helmet}{RGB}{102, 212, 119}
\definecolor{color_wheel}{RGB}{152, 114, 178}
\definecolor{color_shirt}{RGB}{218, 176, 225}
\definecolor{color_pant}{RGB}{114, 129, 157}
\definecolor{color_flow}{RGB}{173, 216, 230}

\newcommand{\ourmodel}{OvSGTR }

\begin{document}
\title{Expanding Scene Graph Boundaries: Fully Open-vocabulary Scene Graph Generation via Visual-Concept Alignment and Retention}

\titlerunning{OvSGTR}

\author{Zuyao Chen\inst{1,2}\orcidlink{0000-0002-7344-1101} \and
Jinlin Wu\inst{2,3}\orcidlink{0000-0001-7877-5728}  \and
Zhen Lei\inst{2,3,4} \orcidlink{0000-0002-0791-189X} \and 
Zhaoxiang Zhang \inst{2, 3,4} \orcidlink{0000-0003-2648-3875} \and 
Chang Wen Chen \inst{1} \orcidlink{0000-0002-6720-234X} \thanks{Corresponding author.} 
}
\authorrunning{Z. Chen et al.}
%
\institute{The Hong Kong Polytechnic University  \\ 
\and 
Centre for Artificial Intelligence and Robotics, HKISI-CAS \and 
State Key Laboratory of Multimodal Artificial Intelligence Systems, Institute of Automation, Chinese Academy of Sciences \and 
School of Artificial Intelligence, University of Chinese Academy of Sciences  
\email{zuyao.chen@connect.polyu.hk} \\ 
\email{\{wujinlin2017, zhen.lei, zhaoxiang.zhang\}@ia.ac.cn} 
\email{changwen.chen@polyu.edu.hk}
}





\maketitle

\begin{abstract}
Scene Graph Generation (SGG) offers a structured representation critical in many computer vision applications.
Traditional SGG approaches, however, are limited by a closed-set assumption, restricting their ability to recognize only predefined object and relation categories. 
To overcome this, we categorize SGG scenarios into four distinct settings based on the node and edge: Closed-set SGG, 
Open Vocabulary (object) Detection-based SGG (OvD-SGG), Open Vocabulary Relation-based SGG (OvR-SGG), and Open Vocabulary Detection + Relation-based SGG (OvD+R-SGG).
While object-centric open vocabulary SGG has been studied recently, 
the more challenging problem of relation-involved open-vocabulary SGG remains relatively unexplored.
To fill this gap,  we propose a unified framework named OvSGTR towards fully open vocabulary SGG from a holistic view. 
The proposed framework is an end-to-end transformer architecture, which learns a visual-concept alignment for both nodes and edges, enabling the model to recognize unseen categories. 
For the more challenging settings of relation-involved open vocabulary SGG, the proposed approach integrates relation-aware pre-training utilizing image-caption data and retains visual-concept alignment through knowledge distillation.
Comprehensive experimental results on the Visual Genome benchmark demonstrate the effectiveness and superiority of the proposed framework.
Our code is available at \url{https://github.com/gpt4vision/OvSGTR/}.
\end{abstract}

\begin{figure}[t]
    \centering
   \includegraphics[height=0.45\textheight]{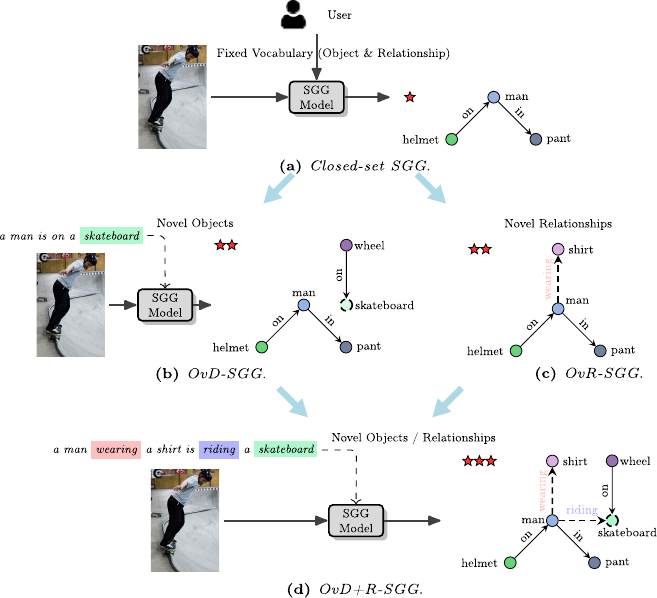}
   \caption{Illustration of SGG Scenarios (best view in color). Dashed nodes or edges in (a) - (d) refer to unseen category instances, and stars refer to the difficulty of each setting. 
    Previous works \cite{xu2017scene, zellers2018neural, tang2019learning, tang2020unbiased, chiou2021recovering, li2021bipartite, chen2019knowledge, zhang2019graphical} mainly focus on \textit{Closed-set SGG} and few studies \cite{he2022towards, zhang2023learning} cover \textit{OvD-SGG}. In this work, we give a more comprehensive study towards fully open vocabulary SGG.
    }
    \label{fig:scenarios}
\end{figure}
\section{Introduction}

Scene Graph Generation (SGG) aims to generate a descriptive graph that localize objects in an image and simultaneously perceive visual relationships among object pairs.  
Such a structured representation has gained much attention, serving as a foundational component in many vision applications, including image captioning \cite{yang2019auto, chen2020say, gu2019unpaired, wang2019role, nguyen2021defense}, visual question answering \cite{teney2017graph, nuthalapati2021lightweight, kenfack2020robotvqa, lee2019visual}, and image generation \cite{johnson2018image, yang2022diffusion}.

Despite significant advancements in SGG, prevailing approaches predominantly operate within a confined set-up, \ie, they constrain object and relation categories to a predefined set. 
This setting hampers the broader applicability of SGG models in diverse real-world applications.
Influenced by the achievements in open vocabulary object detection \cite{zareian2021open, wu2023aligning, li2022grounded, zhong2022regionclip, du2022learning},  recent works\cite{he2022towards, zhang2023learning} attempt to extend the SGG task from closed-set to open vocabulary domain.
However, they focus on an object-centric open vocabulary setting, which only considers the scene graph nodes. 
A holistic approach to open vocabulary SGG requires a comprehensive analysis of nodes and edges. 
This raises two crucial questions that serve as the driving force behind our research:
\textit{\textbf{Can the model predict unseen objects or relationships ?}}
\textit{\textbf{What if the model encounters both unseen objects and unseen relationships?}}

Given these two questions, we recognize the need to re-evaluate the traditional settings of SGG and propose four distinct scenarios: \textit{Closed-set SGG}, Open Vocabulary (object) Detection-based SGG (\textit{OvD-SGG}), which expands to detect objects beyond a closed set, 
Open Vocabulary Relation-based SGG (\textit{OvR-SGG}), focusing on identifying a broader range of object relationships, and 
Open Vocabulary Detection+Relation-based SGG (\textit{OvD+R-SGG}), which combines open vocabulary detection and relation analysis, as shown in \cref{fig:scenarios}.
\textbf{1)} \textit{Closed-set SGG}, extensively studied in previous works \cite{xu2017scene, zellers2018neural, tang2019learning, tang2020unbiased, chiou2021recovering, li2021bipartite, chen2019knowledge, zhang2019graphical}, involves predicting nodes (\ie, objects) and edges (\ie, relationships) from a predefined set.
Generally, \textit{Closed-set SGG} focuses on feature aggregation and unbiased learning for long-tail problems.  
\textbf{2)} \textit{OvD-SGG}, which has recently gained attention \cite{zhang2023learning}, extends \textit{Closed-set SGG} from the node perspective, aiming to recognize unseen object categories during inference. 
However, it still operates on a limited set of relationships.
\textbf{3)} On the other hand, \textit{OvR-SGG} introduces open vocabulary settings from the edge perspective, requiring the model to predict unseen relationships, \textit{a more challenging task due to the absence of pre-trained relation-aware models and the dependence on less accurate scene graph annotations.}
Specifically,  \textit{OvD-SGG} omits all unseen object categories during training, resulting in a graph with fewer nodes but correct edges. By contrast, \textit{OvR-SGG} eliminates all unseen relation categories during training, yielding a graph with fewer edges. As a result, 
\textit{the model for \textit{OvR-SGG} is required to distinguish unseen relationships from ``background''}.
\textbf{4)} The most challenging scenario, \textit{OvD+R-SGG}, involves both unseen objects and unseen relationships, resulting in sparse and less accurate graphs for learning. 
These distinct settings present different intrinsic characteristics and unique challenges.

With a clear understanding of the challenges posed by these settings, we introduce \emph{\ourmodel} (\emph{Open-vocabulary Scene Graph Transformers}), a novel framework designed to address the complexities of open vocabulary SGG. 
Our approach not only predicts unseen objects or relationships but also handles the challenging scenario where both object and relationship categories are unseen during the training phase. 
\textit{\ourmodel} employs a visual-concept alignment strategy for nodes and edges, utilizing image-caption data for weakly-supervised relation-aware pre-training.
The framework comprises three main components: a frozen image backbone for visual feature extraction, a frozen text encoder for textual feature extraction, and a transformer for decoding scene graphs.
During the relation-aware pre-training, the captions are parsed into relation triplets, \ie, (\textit{subject}, \textit{relation}, \textit{object}), which provides a coarse and unlocalized scene graph for supervision.
For the fine-tuning phase, relation triplets with location information (\ie, bounding boxes) are sampled from manual annotations. 
These relation triplets are associated with visual features, and visual-concept similarities are computed for nodes and edges, respectively. 
Predictions regarding object and relation categories are subsequently derived from visual-concept similarities, which promotes the model's generalization ability on unseen object and relation categories.

Upon evaluating the settings for relation-involved open vocabulary SGG (\ie, \textit{OvR-SGG} and \textit{OvD+R-SGG}), we empirically identified a significant issue of catastrophic forgetting pertaining to relation categories.
Catastrophic forgetting leads to a degradation in the model's ability to recall previously learned information from image-caption data when exposed to new SGG data with fine-grained annotations.
To preserve the semantic space while minimizing compromises on the new dataset, we propose visual-concept retention with a knowledge distillation strategy to mitigate this concern. 
The knowledge distillation component utilizes a pre-trained model on image-caption data as a teacher to guide the learning of our student model, ensuring the retention of a rich semantic space of relations. 
Simultaneously, the visual-concept retention ensures that the model maintains its proficiency in recognizing new relations.

In short, the contributions of this work can be summarized as follows, \begin{itemize}
    \item  We give a comprehensive and in-depth study on open vocabulary SGG from the perspective of nodes and edges, discerning four distinct settings including \textit{Closed-set SGG}, \textit{OvD-SGG}, \textit{OvR-SGG}, and \textit{OvD+R-SGG}. Our analysis delves into both quantitative and qualitative aspects, providing a holistic understanding of the challenges associated with each setting;
    \item The proposed framework is fully open vocabulary as both nodes and edges are extendable and flexible to unseen categories, which largely expand the application of SGG models in the real world;
    \item  The integration of a visual-concept alignment  with image-caption data significantly enriches relation-involved open vocabulary SGG, while our visual-concept retention strategy effectively counters catastrophic forgetting; 
    \item  Extensive experimental results on the VG150 benchmark demonstrate the effectiveness of the proposed framework, showcasing state-of-the-art performances across all settings.
\end{itemize}
\section{Related Work}
\textbf{Scene Graph Generation (SGG)} aims to generate an informative graph that localizes objects and describes the relationships between object pairs.
Previous methods mainly focus on contextual information aggregation \cite{xu2017scene, zellers2018neural, tang2019learning} , and unbias learning for long-tail problem \cite{tang2020unbiased, chiou2021recovering, li2021bipartite}. Typically,  a closed-set object detector like Faster-RCNN is used and cannot handle unseen objects or unseen relations, which limits the application of SGG models in the real world. Recent works \cite{he2022towards, zhang2023learning} attempt to extend closed-set SGG to object-centric open vocabulary SGG ; However, they still fail to generalize on unseen relations and the combination of unseen objects and unseen relations.

An alternative approach to boosting the SGG task lies in the utilization of weak supervision, particularly by harnessing image caption data, leading to the emergence of language-supervised SGG \cite{zhong2021learning, li2022integrating, zhang2023learning}. 
This method of language supervision provides a cheaper way for SGG learning than expensive and time-cost manual annotation. 
Although previous research \cite{zhong2021learning, li2022integrating, zhang2023learning} has shown the potential of this technique,  it remains confined predominantly to closed-set relation recognition. 
By contrast, our framework is fully open vocabulary. It discards the synsets matching as used in \cite{zhong2021learning, zhang2023learning}, enabling our model to learn rich semantic concepts for generalization on downstream tasks. 
Furthermore, we also build a connection between language-supervised SGG and open vocabulary SGG, in which language-supervised SGG aims to reduce the alignment gap between visual and language semantic space.

In essence, our work can be perceived as a generalization of open vocabulary SGG, harmoniously integrated with closed-set SGG.  
To our understanding, ours is a pioneering effort in formulating a consolidated framework dedicated to realizing a fully open vocabulary SGG, encompassing both the nodes and edges of scene graphs.

\noindent\textbf{Vision-Language Pretraining (VLP)} has gained increasing attention recently for numerous vision-language tasks. Generally, the core problem of vision-language pretraining is learning an alignment for visual and language semantic space. For instance, CLIP \cite{radford2021learning} shows promising zero-shot image classification capabilities by utilizing contrastive learning on large-scale image-text datasets. Later, many methods \cite{li2022grounded, liu2023grounding, zhong2022regionclip} have been proposed for learning a fine-grained alignment for image region and language data, enabling the object detector to detect unseen objects by leveraging language information. 
The success of VLP on downstream tasks provides an exemplar for learning an alignment between visual features and relation concepts, which is fundamental to building a fully open vocabulary SGG framework.

\noindent\textbf{Open-vocabulary Object Detection (OvD)} expects to detect unseen classes in inference, which breaks the limitation of a fixed pre-defined object set 
(\eg, 80 categories in COCO). 
To accomplish this goal,  Ov-RCNN \cite{zareian2021open} transfers semantic knowledge learned from captions to the downstream object detection task.
It is worth noting that supervision signals for unseen or novel classes are excluded during training detectors, 
while unseen classes can be included in the large vocabulary set of captions. 
Except for OvD, a series of methods and applications have been 
developed such as open-vocabulary segmentation \cite{DBLP:conf/eccv/GhiasiGCL22}, open-vocabulary video understanding \cite{DBLP:journals/corr/abs-2109-08472}, 
and open-vocabulary SGG \cite{he2022towards, zhang2023learning, li2024pixels}. 
A more in-depth analysis of open-vocabulary learning can refer to the literature \cite{wu2024towards}  and \cite{DBLP:journals/corr/abs-2307-09220}. 

\section{Methodology}
\begin{figure}[t]
\centering
\includegraphics[width=0.95\textwidth]{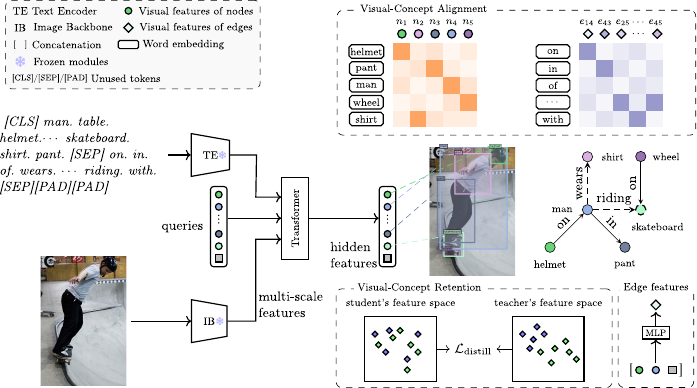}
  \caption{Overview of our proposed \ourmodel. The proposed \ourmodel is equipped with a frozen image backbone to extract visual features, 
  a frozen text encoder to extract text features, and a transformer for decoding scene graphs.
 Visual features for nodes are the output hidden features of the transformer; 
Visual features for edges are obtained via a light-weight relation head (\ie, with only two-layer MLP). 
  Visual-concept alignment associates visual features of nodes/edges with corresponding text features. 
  Visual-concept retention aims to transfer the teacher's capability of recognizing unseen categories to the student.} 
  \label{fig:framework}
\end{figure}

Given an image $I$, the objective of the SGG task is to produce a descriptive graph $\mathcal{G} = (\mathcal{V}, \mathcal{E})$ , in which node $v_i \in \mathcal{V}$ has location information (\ie, bounding box) and object category information, and edge $e_{ij} \in \mathcal{E}$ measure the relationship between node $v_i$ and node $v_j$. 
For open-vocabulary settings,  the label set $\mathcal{C}$ (either for the node or the edge) is split into two disjoint sets : \emph{base classes} $\mathcal{C}_B$ and 
\emph{novel classes} $\mathcal{C}_N$ 
($\mathcal{C}_B \cup \mathcal{C}_N=\mathcal{C}$, $\mathcal{C}_B \cap \mathcal{C}_N=\emptyset$). 

\subsection{Fully Open Vocabulary Architecture}
As shown in \cref{fig:framework}, \textit{\ourmodel} is a DETR-like architecture that comprises three primary components: a visual encoder for image feature extraction, a text encoder for text feature extraction, and a transformer for the dual purposes of object detection and relationship recognition. 
When provided with paired image-text data, \textit{\ourmodel} is adept at generating corresponding scene graphs. 
To ease the optimization burden, the weights of both the image backbone and the text encoder are frozen during training.

\textbf{Feature Extraction.} Given an image-text pair, the model will extract multi-scale visual features with an image backbone like Swin Transformer \cite{liu2021swin} and extract text features via a text encoder like BERT \cite{kenton2019bert}. Visual and text features will be fused and enhanced via cross-attention in the deformable encoder module of the transformer.

\textbf{Prompt Construction.} The text prompt is constructed by concatenating all possible (or sampled) noun phrases and relation categories, \eg, \textit{[CLS] girl. umbrella. table. bathing suit.$\cdots$ zebra. [SEP]  on. in. wears. $\cdots$ walking. [SEP][PAD][PAD]}, which is as similar as GLIP \cite{li2022grounded} or Grounding  DINO \cite{liu2023grounding} concatenating all noun phrases. 
For a large vocabulary set during training, we randomly sample
negative words from the vocabulary set and constrain the number
of positive and negative words to be M (\eg, M=80).

\textbf{Node Representation.}
Given $K$ object queries,  the model follows standard DETR to output $K$ hidden features $\lbrace \bm{v}_i \rbrace_{i=1}^{K}$, which follow a bbox. head to decode the location information (\ie, 4-d vectors), and a cls. head responsible for category classification. The bbox. head is a three-layer fully connected layers.
The cls. head is parameter-free, which computes the similarity between hidden features and text features. 
These hidden features are served as the visual representation for predicted nodes.

\textbf{Edge Representation.}  Contrary to  a complex and heavy message-passing mechanism for obtaining relation features, 
we design a lightweight relation head that concatenates node features for the subject and object, and relation query features. 
To learn a relation-aware representation, we use a random initialized embedding for querying relations. 
This relation-aware embedding will interact with image and text features by cross-attention in the decoder stage. 
Building on this design, given any possible subject-object pair $(s_i, o_j)$,  its edge representation can be obtained with 
$
    \bm{e}_{{s_i} \rightarrow {o_j}} =  f_{\theta}([\bm{v}_{s_i}, \bm{v}_{o_j}, \bm{r}]) 
    \label{eq:rln}
$, 
where $\bm{v}_{s_i}, \bm{v}_{o_j}$ are node representation for the subject and object respectively, $\bm{r}$ refers to the relation query features,  $[\cdot]$ refers to concatenation operation, and $f_{\theta}$ denotes a two-layer multi-perceptrons.

\textbf{Loss Function.}
Following previous DETR-like methods \cite{zhu2021deformable, liu2023grounding} , we use L1 loss and GIoU loss \cite{rezatofighi2019generalized} for bounding box regression. For object or relation classification, we use Focal Loss \cite{lin2017focal} as the contrastive loss between prediction and language tokens. 

To decode object and relation categories in a fully open vocabulary way, the fixed classifier (one fully connected layer) is replaced with a visual-concept alignment, which will be introduced in \cref{sec:lvca}.
\subsection{Learning Visual-Concept Alignment} 
\label{sec:lvca}
Visual-concept alignment associates visual features for nodes or edges with corresponding text features.
For node-level alignment,  take an image as example,   
the model will output $K$ predicted nodes $\lbrace \tilde{v}_i \rbrace_{i=1}^{K}$. 
These predicted nodes  must be matched and aligned with $N$ ground-truth nodes    $\lbrace {{v}_i}\rbrace_{i=1}^{N} $. 
 The matching is formulated as a bipartite graph matching, similar to the approach in standard DETR.
 This can be expressed as
$  
 		\max_{\bm{M}}   \sum_{i=1}^{N} \sum_{j=1}^{K}  \mathrm{sim}(v_i,   \tilde{v}_j) \cdot \bm{M}_{ij}
   \label{eq:bipart}
$.
Here, $\mathrm{sim}(\cdot , \cdot)$ measures the similarity between the predicted node and the ground-truth,  which generally consider both the location (\ie, bounding box) and category information. 
$\bm{M} \in \mathbb{R}^{N \times K}$ is a binary mask where the element $\bm{M}_{ij}=1$ indicates a match between node $v_i$ and node $\tilde{v}_j$. Conversely, a value of 0 indicates no match. 
For any matched pair $(v_i, \tilde{v}_j)$,  we directly maximize its similarity, in which the distance between bounding boxes is determined by the L1 and GIoU losses,   and category similarity is described  as 
\begin{equation}
	\mathrm{sim}_{\mathrm{cat}} (v_i, \tilde{v}_j) = \sigma (<\bm{w}_{v_i}, \bm{v}_j>)
	\label{eq:align1}
\end{equation}
where $\bm{w}_{v_i}$ is the word embedding for node $v_i$, 
$\bm{v}_j$ is the visual representation for predicted node $\tilde{v}_j$, 
$<\cdot, \cdot>$ refers to the dot product of two vectors,
and $\sigma$ refers to the sigmoid function. 
This \cref{eq:align1} seeks to align visual features for nodes with their prototypes in text space. 

To extend relation recognition from closed-set to open vocabulary, one intuitive idea is to learn a visual semantic space in which visual features and text features for relations are aligned.  Specifically,  given a text input $\bm{t}$ and a text encoder $E_t$, a relation feature $\bm{e}$, the alignment score is defined as 
$
    s(\bm{e}) = <\bm{e},  f(E_t(\bm{t})) >
$,
where $f$ is one fully connected layer, and $<\cdot, \cdot>$ refers to the dot product of two vectors. 
Once the alignment score computed, we can calculate a binary cross entropy loss with given ground truths.  The loss can be formulated as 
\begin{equation}
        \mathcal{L}_{\mathrm{bce}} = \frac{1}{|\mathcal{P}| + |\mathcal{N}|} \sum_{\bm{e} \in \mathcal{P}\cup \mathcal{N}} \lbrace -y_{\bm{e}} \log \sigma(s(\bm{e}))  
        - (1-y_{\bm{e}}) \log (1 - \sigma(s(\bm{e}))) \rbrace 
    \label{eq:bce}
\end{equation}
where $\sigma$ refers to sigmoid function, $y_{\bm{e}}$ is a one hot vector where ``1'' index positive tokens, and $\mathcal{P}$, $\mathcal{N}$ refer to positive and negative samples set for relations.

Learning such visual-concept alignment is non-trivial as there is a lack of relation-aware pre-trained models on large-scale datasets. In contrast, object-language alignment can be beneficial from pre-trained models such as CLIP \cite{radford2021learning} and GLIP \cite{li2022grounded}. 
On the other hand, manual annotation of scene graphs is time-consuming and expensive, which makes it hard to obtain large-scale SGG datasets.
To tackle this problem, we leverage image-caption data as a weak supervision for relation-aware pre-training. 
Specifically, given an image-caption pair without bounding boxes annotation, we utilize an off-the-shelf language parser \cite{mao2018parser} to parse relation triplets from the caption. 
These relation triplets are associated with predicted nodes by optimizing \cref{eq:bipart}, and only triplets with high confidence (\eg, object score is greater than 0.25 for both subject and object) are reserved in scene graphs as pseudo labels. 
Utilizing these pseudo labels as a form of weak supervision, the model is enabled to learn rich concepts for objects and relations with image-caption data. 
\subsection{Visual-Concept Retention with Knowledge Distillation}
\label{sec:vcp}
Through learning a visual-concept alignment as described in \cref{sec:lvca}, the model is expected to recognize rich objects and relations beyond a fixed small set. 
However, we empirically find that directly optimizing the model by \cref{eq:bce} on a new dataset will meet catastrophic forgetting even if we have a relation-aware pre-trained model.
On the other hand, in \textit{OvR-SGG} or \textit{OvD+R-SGG} settings,  unseen (or novel) relationships are removed from the graph,  which increases the difficulty as the model is required to distinguish novel relations from ``background''.  
To mitigate this problem,  we adopt a knowledge distillation strategy to maintain the consistency of learned semantic space. 
Specifically, we use the initialized model pre-trained on image caption data as the teacher. The teacher has learned a rich semantic space for relations, \eg, there exist ${\sim}2.5k$ relation categories parsed from COCO caption \cite{chen2015microsoft} data. 
The student's edge features should be as close as the teacher's for the same negative samples. 
Thus, the loss for relationship recognition can be formulated as 
\begin{equation}
       \mathcal{L}_{\mathrm{distill} } = \frac{1}{|\mathcal{N}|} \sum_{\bm{e} \in \mathcal{N}} || \bm{e}^s - \bm{e}^t||_1 
    \label{eq:distill}
\end{equation}
where $\bm{e}^s$ and $\bm{e}^t$ refer to the student's and teacher's edge features, respectively. 
The total loss is given as 
$
    \mathcal{L} = \mathcal{L}_{\mathrm{bce}} + \lambda \mathcal{L}_{\mathrm{distill}}
$, 
where $\lambda$  controls the ratio of ground truths supervision and distillation part.

\section{Experiments}   
\subsection{Datasets and Experiment setup}
\textbf{Datasets.}
The widely used VG150 dataset \cite{xu2017scene} contains $150$ object and $50$ relation categories annotated by humans.
Of its $108, 777$ images,  $70\%$   are used for training, $5,000$  for validation, and the rest  for testing. 
Following $\text{VS}^3$ \cite{zhang2023learning}, we exclude images used in pre-trained object detector Grounding DINO \cite{liu2023grounding} , retaining $14,700$ test images. 
And  we use an off-the-shelf language parser \cite{mao2018parser} to parse relation triplets from image caption, which yields ${\sim}117k$ images with ${\sim}44k$ phrases and ${\sim}2.5 k$ relations for COCO caption training set. 
To showcase the scalability of our model, we concat COCO caption data \cite{chen2015microsoft}, Flickr30k \cite{plummer2015flickr30k}, and SBU Captions \cite{ordonez2011im2text} to construct a large-scale dataset for scene graph pre-training, resulting in ${\sim}569k$ images with ${\sim}198k$ type phrases and ${\sim}5k$ relations. 

\noindent\textbf{Metrics.}
We adopt the \textbf{SGDET} \cite{xu2017scene, tang2020unbiased} protocol (alias: \textbf{SGGen})  for fair comparison and report the performance on Recall@K (K=20/50/100) for each settings. 
Mean Recall@K (mR@K, K=20/50/100) and inference speed 
are reported under the setting of \textit{Closed-set SGG}.

\noindent\textbf{Implementation details.}
We use pre-trained Grounding DINO \cite{liu2023grounding} models  to initialize our model, and keep the visual backbone (\ie, Swin-T or Swin-B) and text encoder (\ie, BERT-base \cite{kenton2019bert}) as frozen. Other modules like relation-aware embedding are initialized randomly. 
And $100$ object detections per image are selected for pairwise relation recognition. 
 
\subsection{Compared with State-of-the-arts}
\begin{table}[t]
    \centering
    \caption{Experimental results of \textit{Closed-set SGG} on VG150 test set. ``40M/177M'' in Params. refers to 40M trainable parameters and 177M total parameters.
    Inference time is benchmarked on an NVIDIA RTX 3090 GPU with batch size 1 and an input resolution $1000\times 600$.
    Time for \text{SGNLS} \cite{zhong2021learning}  is benchmarked on an NVIDIA A100 GPU (80G) due to memory out of usage.
    }    
    \resizebox{0.95\textwidth}{!}{
    \begin{tabular}{l|c|c|c|>{\centering\arraybackslash}p{8mm}>{\centering\arraybackslash}p{8mm}>{\centering\arraybackslash}p{8mm}
    |>{\centering\arraybackslash}p{7mm}>{\centering\arraybackslash}p{7mm}>{\centering\arraybackslash}p{7mm}|c}
   \toprule
         SGG model& Backbone & Detector& Params. & 
         \multicolumn{3}{c|}{R@20/50/100} & 
         \multicolumn{3}{c|}{mR@20/50/100} &  Time (s)   \\ 
        \midrule 
          IMP \cite{xu2017scene} &  RX-101  & & 146M/308M &  17.7 & 25.5  & 30.7 &2.7  &4.1 &5.3  &  0.25\\
         MOTIFS \cite{zellers2018neural} & RX-101&Faster &205M/367M & 25.5 & 32.8 & 37.2&   5.0& 6.8& 7.9 &   0.27\\
         VCTREE \cite{tang2019learning} &   RX-101& R-CNN &197M/358M & 24.7 & 31.5 & 36.2&  - & - & - & 0.38\\
         \text{SGNLS} \cite{zhong2021learning} &   RX-101&  &  165M/327M  & 24.6 & 31.8 & 36.3&  -& -& -  & > 7\\
         HL-Net \cite{lin2022hl} &    RX-101& & 220M/382M& 26.0 & 33.7 & 38.1& -  &- &-  & 0.10 \\
        \hline
         FCSGG \cite{liu2021fully} &   HRNetW48& - & 87M/87M  &13.6 & 18.6 & 22.5&  2.3 & 3.2& 3.9 &  0.13\\
         SGTR \cite{li2022sgtr} &   R-101& DETR &  36M/96M & - &24.6 & 28.4& - & -&  -&   0.21\\
         $\text{VS}^{3}$ \cite{zhang2023learning} &   Swin-T& -& 93M/233M&26.1 & 34.5 & 39.2& -
          & -& -  & 0.16\\
         $\text{VS}^{3}$ \cite{zhang2023learning} &   Swin-L& -  & 124M/432M & 27.3 & 36.0 & 40.9&  4.4 &6.5 &7.8  & 0.24 \\       
         \hline 
         \rowcolor{lightgray}
         \ourmodel &   Swin-T& DETR & 
          41M/178M
         &\textbf{27.0} & \textbf{35.8} & \textbf{41.3}& 
         \textbf{5.0} &\textbf{7.2} & \textbf{8.8} & 0.13\\
         \rowcolor{lightgray}
         \ourmodel &   Swin-B& DETR & 
         41M/238M
         & \textbf{27.8} & \textbf{36.4} & \textbf{42.4} & \textbf{5.2} & \textbf{7.4}&  \textbf{9.0} &  0.19 \\
         \bottomrule 
    \end{tabular} }
    \label{tab:closed}
\end{table}
\textbf{Closed-set SGG Benchmark.} The \textit{Closed-set SGG} setting follows previous works \cite{zellers2018neural, tang2020unbiased, li2022sgtr, xu2017scene, zhang2023learning}, utilizing the VG150 dataset \cite{xu2017scene} with full manual annotations for training and evaluation.
Experimental results on the VG150 test set are reported in \cref{tab:closed}, demonstrating that the proposed model outperforms all competitors. 
Notably, when compared to the recent $\text{VS}^3$ \cite{zhang2023learning}, \textit{\ourmodel}  (\textit{w.} Swin-T) shows a performance gain of up to $3.8\%$ for R@50 and $5.4\%$ for R@100. 
The performance gain regarding mR@K reflects that our model handles the long-tail bias better than others.

Moreover, while many previous works rely on a complex message-passing mechanism to extract relation features, our model achieves strong performance with a simpler relation head consisting of only two MLP layers.
For example, \textit{\ourmodel}  (\textit{w.} Swin-T) achieves a comparable, even better result than $\text{VS}^3$ (\textit{w.} Swin-L). At the same time, our model has fewer trainable parameters (41M vs. 124M) and lower inference latency (0.13 s vs. 0.24 s).
\begin{table}[t]
    \centering
    \caption{Experimental results (R@50/100) of \textit{OvD-SGG} setting on VG150 test set.
    Following $\text{VS}^3$ \cite{zhang2023learning},
    OvSGTR chooses image regions that best match the ground-truth objects in post-processing for PREDCLS. 
    }    
    \resizebox{0.8\textwidth}{!}{
    \begin{tabular}{l|>{\centering\arraybackslash}p{24mm}>{\centering\arraybackslash}p{24mm}|>{\centering\arraybackslash}p{24mm}>{\centering\arraybackslash}p{24mm}
    } 
            \toprule
            \multirow{2}{*}{Method} & \multicolumn{2}{c|}{Base+Novel (Object)} & 
           \multicolumn{2}{c}{Novel (Object)}  \\
              & PREDCLS &  SGDET &   PREDCLS & SGDET \\ 
          \midrule 
            IMP \cite{xu2017scene} & 40.02 / 43.40 & 2.85 / 3.43 & 37.01 / 39.46 & 0.00 / 0.00 \\ 
            MOTIFS \cite{zellers2018neural} & 41.14 / 44.70 & 3.35 / 3.86 & 39.53 / 41.14 & 0.00 / 0.00 \\
            VCTREE \cite{tang2019learning} & 42.56 / 45.84 & 3.56 / 4.05 & 41.27 / 42.52 & 0.00 / 0.00 \\
            TDE \cite{tang2020unbiased} & 38.29 / 40.38 & 3.50 / 4.07 & 34.15 / 36.37& 0.00 / 0.00 \\
            GCA\cite{knyazev2021generative} & 43.48 / 46.26 & - & 42.56 / 43.18 & - \\
            EBM \cite{suhail2021energy} & 44.09 / 46.95 & - & 43.27/44.03 & - \\
            SVRP \cite{he2022towards} & 47.62 / 49.94 & - & 45.75 / 48.39 & - \\
            $\text{VS}^3$ \cite{zhang2023learning} (\footnotesize Swin-T)& 50.10 / 52.05 & 15.07 / 18.73 &
            46.91 / 49.13 & 10.08 / 13.65 \\    
            \hline 
            \rowcolor{lightgray}
            \ourmodel (\footnotesize Swin-T) & \textbf{60.58 / 62.10} & \textbf{18.14 / 23.20} & \textbf{59.01 / 60.65}& \textbf{12.06 / 16.49} \\
            \rowcolor{lightgray}
           \ourmodel (\footnotesize Swin-B) & \textbf{60.83 / 62.33} &  \textbf{21.35 / 26.22}   & \textbf{59.30 / 60.95}&   \textbf{15.58 / 19.96} \\ 
          \bottomrule
    \end{tabular}}
    \label{tab:ovd}
\end{table}
\begin{table}[t]
    \centering
    \caption{Experimental results of \textit{OvR-SGG} setting on VG150 test set. 
     $\dag$ refers to \textit{w.o.} distillation.
    }    
    \resizebox{0.8\textwidth}{!}{
    \begin{tabular}{l|cc|cc} 
            \toprule
            \multirow{2}{*}{Method}  &\multicolumn{2}{c|}{Base+Novel (Relation)} & 
           \multicolumn{2}{c}{Novel (Relation)}  \\
             & R@50  & R@100  &    R@50 & R@100  \\ 
          \midrule 
          IMP \cite{xu2017scene} & 12.56    &  14.65 & 0.00 & 0.00 \\
          MOTIFS \cite{zellers2018neural} & 15.41 & 16.96& 0.00& 0.00 \\
          VCTREE \cite{tang2019learning} & 15.61 & 17.26 & 0.00 & 0.00  \\
          TDE \cite{tang2020unbiased} & 15.50 & 17.37 &0.00 & 0.00 \\ 
        $\text{VS}^3$ \cite{zhang2023learning} (\footnotesize Swin-T) &
          15.60 & 17.30 & 0.00 & 0.00 \\ 
        \hline 
        \rowcolor{lightgray}
        ${\text{OvSGTR}_{Swin-T}}^\dag$  &   17.71 & 20.00 & 0.34 & 0.41\\ 
        \rowcolor{lightgray}
        $\text{OvSGTR}_{{Swin-T}}$  &   \textbf{20.46} & \textbf{23.86} & \textbf{13.45} & \textbf{16.19} \\ 
        \hdashline 
        \rowcolor{lightgray}
        ${\text{OvSGTR}_{Swin-B}}^\dag$  &   18.58 & 20.84   & 0.08 & 0.10  \\ 
        \rowcolor{lightgray}
        $\text{OvSGTR}_{{Swin-B}}$ & \textbf{22.89}   & \textbf{26.65}  & \textbf{16.39}  & \textbf{19.72}  \\        
          \bottomrule
    \end{tabular}}
    \label{tab:ovr}
\end{table}
\begin{table}[t]
    \centering
    \caption{Experimental results of \textit{OvD+R-SGG} setting on VG150 test set. 
    $\dag$ refers to \textit{w.o.} distillation.
    }    
    \resizebox{0.75\textwidth}{!}
    {
    \begin{tabular}{l|cc|cc|cc} 
            \toprule
            \multirow{2}{*}{Method}  &\multicolumn{2}{c|}{Joint Base+Novel} & 
           \multicolumn{2}{c}{Novel (Object)}  & \multicolumn{2}{c}{Novel (Relation)}  \\
              & R@50  & R@100  &    R@50  & R@100  &    R@50 & R@100 \\ 
          \midrule 
          IMP \cite{xu2017scene} &   0.77  &  0.94 & 0.00 & 0.00 & 0.00 & 0.00 \\
          MOTIFS \cite{zellers2018neural} & 1.00 & 1.12& 0.00 & 0.00 & 0.00 & 0.00 \\
          VCTREE \cite{tang2019learning} & 1.04 & 1.17 & 0.00 &0.00 & 0.00 & 0.00\\
          TDE \cite{tang2020unbiased} & 1.00 & 1.15& 0.00 & 0.00 & 0.00 & 0.00 \\         
        $\text{VS}^3$ \cite{zhang2023learning} (\footnotesize Swin-T) &   5.88 & 7.20 & 6.00 & 7.51 & 0.00 & 0.00 \\ 
        \hline 
        \rowcolor{lightgray}
         ${\text{OvSGTR}_{Swin-T}}^\dag$ &  7.88   &  10.06  & 6.82  &9.23   &0.00 &0.00 \\ 
        \rowcolor{lightgray}
        $\text{OvSGTR}_{{Swin-T}}$ &   \textbf{13.53}   & \textbf{16.36} & \textbf{14.37}  & \textbf{17.44}  & \textbf{9.20} & \textbf{11.19}\\ 
        \hdashline
        \rowcolor{lightgray}
        ${\text{OvSGTR}_{Swin-B}}^\dag$ &   11.23 & 14.21     &  13.27  &  16.83  & 1.78 & 2.57 \\ 
        \rowcolor{lightgray}
        $\text{OvSGTR}_{{Swin-B}}$ &  \textbf{17.11}  &\textbf{21.02}   &\textbf{17.58}    & \textbf{21.72}   & \textbf{14.56}  & \textbf{18.20}  \\         
          \bottomrule
    \end{tabular}}
    \label{tab:ovdr}
\end{table}

\textbf{OvD-SGG Benchmark.} Following previous works \cite{he2022towards, zhang2023learning}, the \textit{OvD-SGG} setting requires the model cannot see novel object categories during training. 
Specifically, $70\%$ selected object categories of VG150 are regarded as base categories, and the remaining $30\%$ object categories are acted as novel categories.  
The experiments under this setting are as same as \textit{Closed-set SGG} except that novel object categories are removed in labels.
After excluding unseen object nodes, the training set of VG150 contains $50,107$ images. 
We report the performance of \textit{OvD-SGG} setting in terms of ``Base+Novel (Object)'' and `` Novel (Object)'' in \cref{tab:ovd}. 
It can be found that the proposed model significantly excel previous methods. 
Compared to $\text{VS}^3$\cite{zhang2023learning}, the performance gain on novel categories is up to $19.6\%$ / $20.8\%$ for R@50 / R@100, which demonstrate the proposed model has more powerful open vocabulary-aware and generalization ability. 
Since the \textit{OvD-SGG} setting only removes nodes with novel object categories, learning process of relations will not be affected; 
This indicates that the performance is more dependent on the open-vocabulary ability of an object detector.

\textbf{OvR-SGG Benchmark.} Different from \textit{OvD-SGG} which removes all unseen nodes , \textit{OvR-SGG} only removes all unseen edges but keep original nodes. 
Considering VG150 has $50$ relation categories, we randomly select $15$ of them as unseen (novel) relation categories. During training, only base relation annotation is available. 
After removing unseen edges, there exists $44,333$ images of VG150 for training.
 Similar as \textit{OvD-SGG}, \cref{tab:ovr} reports the performance of $\textit{OvR-SGG}$ in terms of ``Base+Novel (Relation)'' and ``Novel (Relation)''. 
From \cref{tab:ovr}, the proposed \textit{\ourmodel} notably outperforms other competitors even without distillation. 
However, a marked decline in performance is observed across all techniques, inclusive of \textit{\ourmodel} without distillation, within the ``Novel (Relation)'' categories, underscoring the intrinsic difficulties associated with discerning novel relations in the \textit{OvR-SGG} paradigm.
Nevertheless, with  visual-concept retention, 
the performance of \ourmodel (\textit{w.} Swin-T) on novel relations has been significantly improved from 0.34 (R@50) to 13.45 (R@50).

\textbf{OvD+R-SGG Benchmark.}
This benchmark augments the SGG from a closed-set setting to a fully open vocabulary domain, where both novel object and relation categories are omitted during the training phase. 
For its construction, we combine the split of \textit{OvD-SGG} and \textit{OvR-SGG} and use their base object categories and base relation categories, resulting in $36,425$ images of VG150 for training. 
We report the performance of \textit{OvD+R-SGG} in \cref{tab:ovdr} regarding ``Joint Base+Novel'' (\ie, all object and relation categories considered), ``Novel (Object)'' (\ie, only novel object categories considered), and ``Novel (Relation)'' (\ie, only relation categories considered). 
From \cref{tab:ovdr}, the catastrophic forgetting still occurred in \textit{OvD+R-SGG} as same as \textit{OvR-SGG}, which is alleviated by visual-concept retention in a significant degree. 
When juxtaposed with other methods, our model achieves significant performance gain on all metrics.

\textbf{Overall Analysis.}  
Experimental results present distinct challenges and difficulties in these four settings. Based on these experiments, 
\textbf{1)} many previous methods rely on a two-stage object detector, Faster R-CNN \cite{ren2015faster},  and complicated message-passing mechanism. 
Nevertheless, our model showcases that a one-stage DETR-based framework can significantly surpass R-CNN-like architecture even with only one MLP to obtain feature representation for relations.
\textbf{2)} previous methods with a closed-set object detector struggle to discern objects without textual information under the object-involved open vocabulary SGG (\ie, \textit{OvD-SGG} and \textit{OvD+R-SGG}).
\textbf{3)} the performance drop compared to previous settings reveals that \textit{OvD+R-SGG} is much more challenging than others,  
indicating much room for extensive exploration toward fully open vocabulary SGG.

\subsection{Ablation Study}
\begin{figure}[t]
\centering
\includegraphics[width=0.5\textwidth]{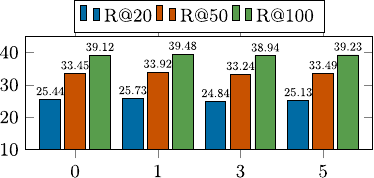}
\caption{Ablation study of  relation queries on VG150 validation set (\textit{Closed-set SGG}).
}
\label{fig:ablation1}
\end{figure} 

\textbf{Effect of Relation Queries.}
We first consider remove relation query embedding. The relation feature is given by $\bm{e}_{{s_i}\rightarrow {o_j}} = f_{\theta}([\bm{v}_{s_i}, \bm{v}_{o_j}]) $,
which only encodes hidden features for the subject and object node.
Further, we extend the \cref{eq:rln} to a more general form as 
$
    \bm{e}_{{s_i}\rightarrow {o_j}} = \frac{1}{M} \sum_{n=1}^M f_{\theta}([\bm{v}_{s_i}, \bm{v}_{o_j}, \bm{r}_n]) 
    \label{eq:avg@rln}
$, 
which averages multiple relation query results. 
As shown in \cref{fig:ablation1}, the model achieves the best performance when the number of relation queries is set to 1. 
This can be interpreted from two aspects. On the one hand, the relation queries interact with all edges during training, which captures global information for the whole dataset. On the other hand, increasing the number of relation-aware queries does not introduce specific supervision yet heavy the optimization burden.

\noindent\textbf{Relation-aware Pre-training.}
\begin{table}[t]
    \centering
    \caption{Comparison with others trained on image captions
    (which is referred to as \textit{language-supervised SGG} in \cite{zhang2023learning}). All models are trained on image-caption data and test on VG150 test set directly. 
    Our models trained on COCO Captions are used as the pre-training models for \textit{OvR-SGG} and \textit{OvD+R-SGG} settings. 
    }     
    \resizebox{0.64\textwidth}{!}
    {
    \begin{tabular}{l|cc|ccc}
    \toprule
         SGG model&  Training Data & Grounding & R@20  & R@50  & R@100  \\
    \midrule
         LSWS  \cite{yelinguistic}& COCO & - & - & 3.28 & 3.69 \\ 
         MOTIFS\cite{zellers2018neural}& COCO & Li \etal  \cite{li2022integrating} & 5.02 & 6.40 & 7.33 \\
         Uniter \cite{chen2020uniter}& COCO  & SGNLS \cite{zhong2021learning} & - & 5.80 & 6.70 \\
         Uniter \cite{chen2020uniter} & COCO & Li \etal  \cite{li2022integrating} & 5.42 & 6.74 & 7.62 \\
         $\text{VS}^{3}$ \cite{zhang2023learning} (\footnotesize Swin-T) & COCO  & GLIP-L \cite{li2022grounded}  & 5.59 & 7.30 & 8.62 \\
         $\text{VS}^3$  \cite{zhang2023learning} (\footnotesize Swin-L) & COCO & GLIP-L \cite{li2022grounded}  & 6.04 & 8.15 & 9.90 \\ 
    \hline
    $\text{VS}^3$  \cite{zhang2023learning} (\footnotesize Swin-L)
    & VG Caption & GLIP-L \cite{li2022grounded} & 
    10.98 & 15.51 & 19.75 \\
    Ours (Swin-B) & VG Caption & GLIP-L \cite{li2022grounded} &
    16.36 & 22.14  & 26.20 \\ \hline 
    \rowcolor{lightgray}
          &  & Grounding  &   &  &  \\ 
        \rowcolor{lightgray}
        \multirow{-2}{*}{Ours (\footnotesize Swin-T)}& \multirow{-2}{*}{COCO}& DINO-B \cite{liu2023grounding} & \multirow{-2}{*}{\textbf{6.61}} &\multirow{-2}{*}{\textbf{8.92}} &\multirow{-2}{*}{\textbf{10.90}} \\ 
        \rowcolor{lightgray}
         &   & Grounding  &   
          &    &      \\ 
           \rowcolor{lightgray}
        \multirow{-2}{*}{Ours (\footnotesize Swin-B)} & \multirow{-2}{*}{COCO}& DINO-B \cite{liu2023grounding} & \multirow{-2}{*}{\textbf{6.88}} & \multirow{-2}{*}{\textbf{9.30}} &\multirow{-2}{*}{\textbf{11.48}} \\
      \hdashline  
      \rowcolor{lightgray}
          & COCO+ & {Grounding} &  &  &  \\ 
         \rowcolor{lightgray}
          \multirow{-2}{*}{Ours (\footnotesize Swin-T)} & Flickr30k+SBU & DINO-B \cite{liu2023grounding} &\multirow{-2}{*}{\textbf{7.01}} & \multirow{-2}{*}{\textbf{9.43}}& \multirow{-2}{*}{\textbf{11.43}}\\ 
    \bottomrule
    \end{tabular}}
    \label{tab:lang}
\end{table}
We compare \textit{\ourmodel} trained on image caption data with others in \cref{tab:lang}. 
From the result,  the \textit{\ourmodel}  (\textit{w.} Swin-T) with COCO captions outperforms others,  scoring 6.61, 8.92, and 10.90 for R@20, R@50, and R@100, respectively.
When integrated with COCO Captions \cite{chen2015microsoft}, Flickr30k \cite{plummer2015flickr30k}, and SBU Captions \cite{ordonez2011im2text} , its performance peaks at 7.01, 9.43, and 11.43 for the respective metrics. 
The results clearly indicate the effectiveness of the proposed method, particularly when using the more lightweight Swin-B backbone compared to Swin-L; For reference, the zero-shot performance on COCO validation set of GLIP-L \cite{li2022grounded} (\textit{w.} Swin-L) and Grounding DINO-B (\textit{w.} Swin-B) \cite{liu2023grounding} stands at 49.8 AP and 48.4 AP respectively.

\noindent\textbf{Hyper-parameter $\lambda$ for Distillation.}
\cref{tab:lambda} illustrates the impact of varying hyper-parameter $\lambda$ . From the results, when $\lambda=0.1$, the model with distillation achieves the best performance. By contrast, without distillation, a significant decline in performance for novel categories exists, showing the model struggles to retain knowledge inherited from pre-trained models for novel categories. 
\begin{table}[t]
    \centering
    \caption{Impact of hyper-parameter $\lambda$ for distillation loss on VG150 validation set under the setting of \textit{OvR-SGG}.
    $a \rightarrow b$ refers to the performance shift from $a$ (initial checkpoint's performance) to $b$ during training. }    
    \resizebox{0.64\textwidth}{!}
    {
    \begin{tabular}{c|cc|cc}
        \toprule
         $\lambda$  & \multicolumn{2}{c|}{Base+Novel} &
          \multicolumn{2}{c}{Novel} \\
          &  R@50 &  R@100 & R@50 &  R@100 \\
        \midrule
         0 & 7.25 $\rightarrow$ 13.74  & 8.98  $\rightarrow$ 16.11
         & 10.78 $\rightarrow$ 0.32 & 13.24 $\rightarrow$ 0.38 \\ 
         0.1 &   7.25 $\rightarrow$ \textbf{16.00}  & 8.98  $\rightarrow$ \textbf{19.20}
         & 10.78 $\rightarrow$ \textbf{11.54} & 13.24 $\rightarrow$ \textbf{13.94} \\ 
         0.3 &     7.25 $\rightarrow$ 14.35  & 8.98 $\rightarrow$ 17.04  &
         10.78 $\rightarrow$  10.71 & 13.24 $\rightarrow$  12.71
         \\
        0.5 &     7.25 $\rightarrow$ 13.34 & 8.98 $\rightarrow$ 16.08 &
         10.78 $\rightarrow$ 10.90 & 13.24 $\rightarrow$ 13.22 
         \\
       \bottomrule
    \end{tabular}}
    \label{tab:lambda}
\end{table}

\begin{figure}
\centering
\includegraphics[width=0.95\textwidth]{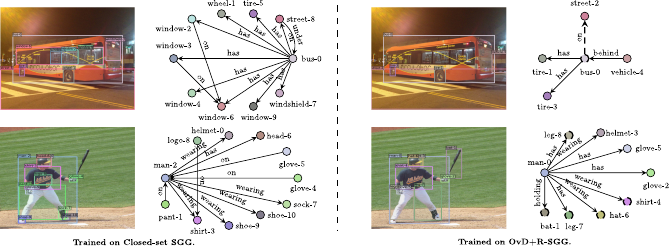}
    \caption{Qualitative results of our model on VG150 test set (best view in color). For clarity, we only show triplets with high confidence in top-20 predictions. Dashed nodes or arrows refer to novel object categories or novel relationships. 
    }
    \label{fig:qual}
\end{figure}
\subsection{Visualization and Discussion}
We present qualitative results of our model trained under \textit{OvD+R-SGG} setting as well as \textit{Closed-set SGG} setting, as shown in \cref{fig:qual}.
From the figure,  the model trained on \textit{Closed-set SGG} tends to generate more dense scene graphs as the whole object and relationship categories are available during training. 
Despite lacking full supervision of novel categories,
the model trained on \textit{OvD+R-SGG} still can recognize novel objects like ``bus'', ``bat'' (which does not exist in VG150 dataset), and novel relationship like ``on'.

\noindent\textbf{Limitations \& Future works.}
One latent limitation of this work is that we utilize an off-the-shelf language parser \cite{mao2018parser} to parse triplets from the caption.
The accuracy of the parser will have a significant impact on the pre-training phase. 
Recently, LLM (large language model) has gained much attention. 
The naive parser can be replaced with a LLM to provide more accurate triplets. 
Moreover,  it is worth discussing 
\textit{Can LLMs benefit the SGG task with fewer manual
annotations?}
or \textit{Can structured representations like scene
graphs benefit for LLMs to alleviate hallucination?}
In the future, we will try to answer these two questions.

\section{Conclusion} 
This work advances the SGG task from a closed set to a fully open vocabulary setting based on the node and edge properties, 
categorizing SGG scenarios into four distinct settings including \textit{Closed-SGG}, \textit{OvD-SGG}, \textit{OvR-SGG}, and \textit{OvD+R-SGG}. 
Towards fully open vocabulary SGG, we design a unified framework named \textit{\ourmodel} with transformers.  
The proposed framework learns to align visual features and concept information with not only base objects,  but also relation categories and generalize on both novel object and relation categories. 
To obtain a transferable representation for relations,
we utilize image-caption data as a weak supervision for relation-aware pre-training. 
In addition, 
visual-concept retention via knowledge distillation is adopted for alleviating the catastrophic forgetting problem in relation-involved open vocabulary SGG. 
We conduct extensive experiments on the VG150 benchmark dataset and have set up new state-of-the-art performances for all settings.

\section*{Acknowledgements}
This research was supported in part by the Hong Kong Research Grants Council (GRF-15229423), 
the Chinese National Natural Science Foundation Projects (U23B2054, 62306313), 
and the InnoHK program.

%
%
\bibliographystyle{splncs04}
\bibliography{egbib}
\end{document}